\pdfoutput=1
\documentclass[10pt,letterpaper,english,british,runningheads,english]{llncs}
\usepackage[T1]{fontenc}
\usepackage[latin9]{inputenc}
\usepackage{color}
\usepackage{amssymb}
\usepackage{graphicx}

\makeatletter

\providecommand{\tabularnewline}{\\}

\usepackage{graphicx}
\usepackage{amsmath,amssymb} 
\usepackage{color}
\usepackage{array}
\usepackage{multirow}
\usepackage{amsmath}
\usepackage{graphicx}
\usepackage{color}
\usepackage[width=122mm,left=12mm,paperwidth=146mm,height=193mm,top=12mm,paperheight=217mm]{geometry}
\usepackage{url}

\pdfpageheight\paperheight
\pdfpagewidth\paperwidth

\providecommand{\tabularnewline}{\\}

\makeatother

\usepackage{babel}
\begin{document}
\pagestyle{headings}
\mainmatter 

\title{Transductive Multi-class and Multi-label Zero-shot Learning}
\titlerunning{}
\authorrunning{Y. Fu, Y. Yang, T. Hospedales, T. Xiang,  S. Gong}
\author{\small{Yanwei Fu, Yongxin Yang, Timothy M. Hospedales, \\Tao Xiang, Shaogang Gong }}
\institute{\small{School of EECS, Queen Mary University of London, UK \\
Email:\{y.fu,yongxin.yang, t.hospedales,  t.xiang, s.gong\}@qmul.ac.uk }} 
\maketitle

\begin{figure}
\begin{centering}
\begin{tabular}{cc}
\includegraphics[scale=0.17]{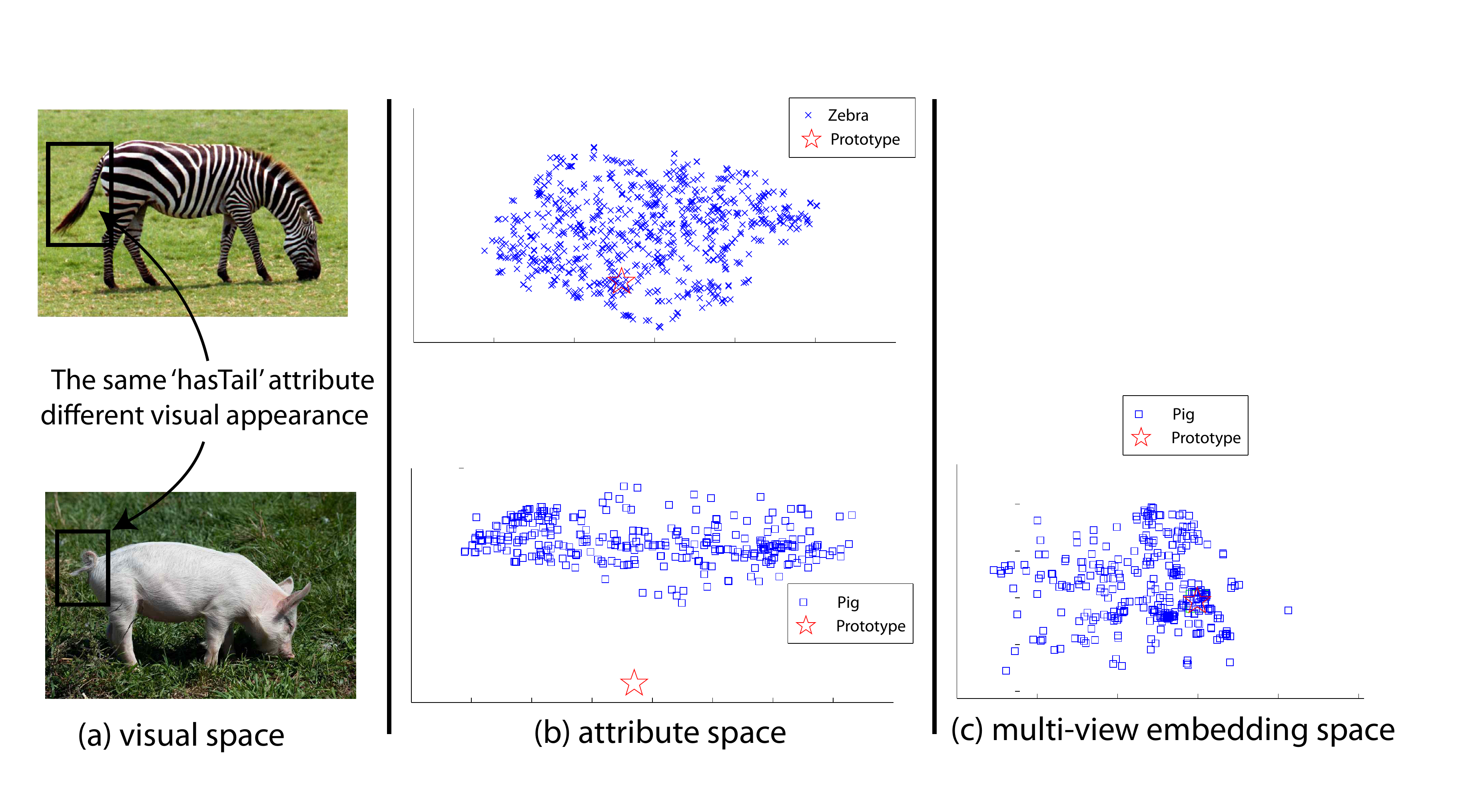} & \includegraphics[scale=0.15]{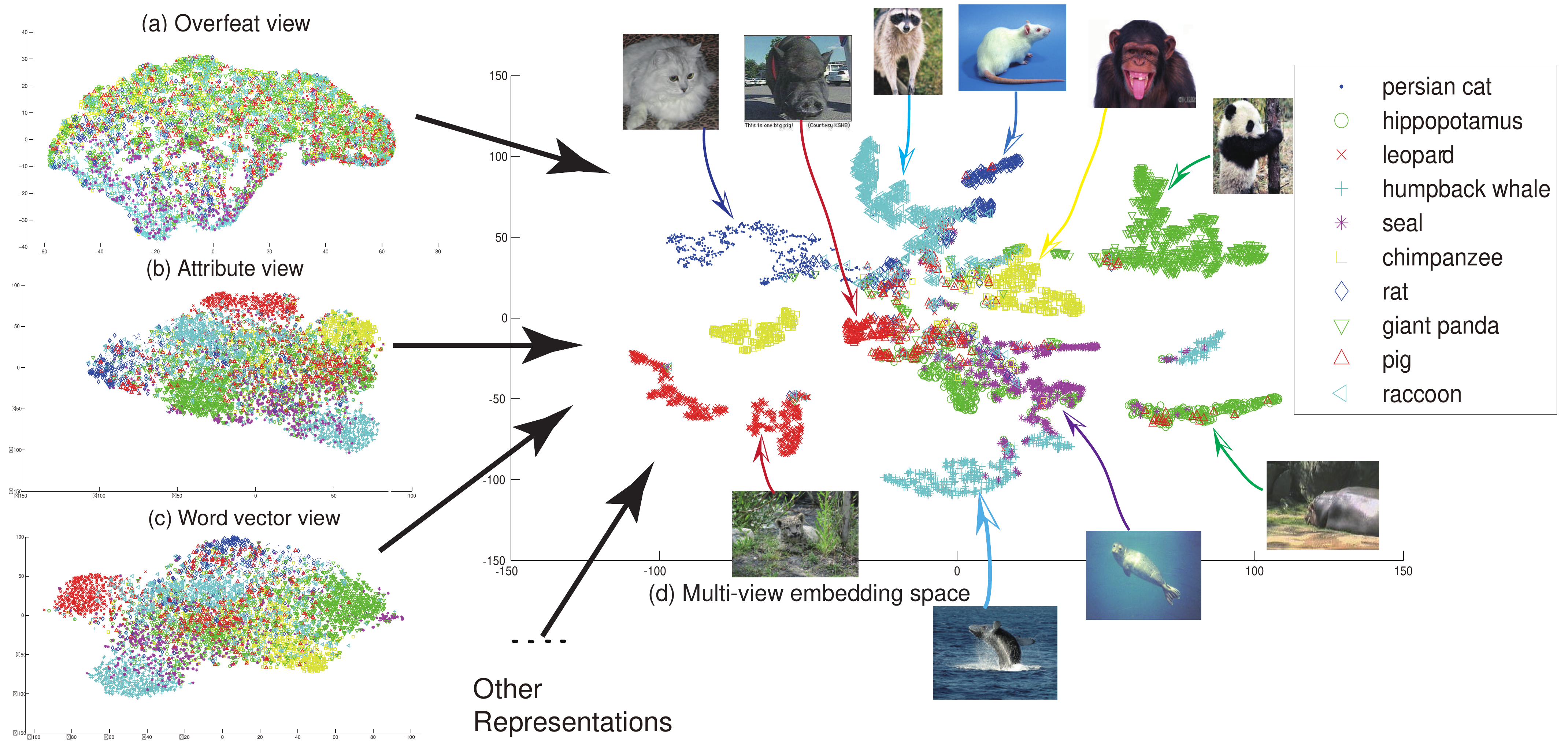}\tabularnewline
\multicolumn{1}{c}{{\tiny{}(A) zero-shot prototypes: red stars;}} & {\tiny{}(B) (a) overfeat view (b) attribute view (c) word vector view}\tabularnewline
{\tiny{}image feature projections: blue} & {\tiny{}(d) multi-view embedding space }\tabularnewline
\end{tabular}
\par\end{centering}

\caption{\scriptsize \label{fig:project_domain_shift_tsne}t-SNE visualisation of AwA: (A) Projection domain shift problem; (B)The instance distance measured by TMV-HLP in our embedding space.}
\end{figure}

Recently, zero-shot learning (ZSL) has received increasing interest.
The key idea underpinning existing ZSL approaches is to exploit knowledge
transfer via an intermediate-level semantic representation which is
assumed to be shared between the auxiliary and target datasets, and
is used to bridge between these domains for knowledge transfer. The
semantic representation used in existing approaches varies from visual
attributes \cite{lampert2009zeroshot_dat,farhadi2009attrib_describe,liu2011action_attrib,yanweiPAMIlatentattrib}
to semantic word vectors \cite{DeviseNIPS13,RichardNIPS13} and semantic
relatedness \cite{RohrbachCVPR12}. However, the overall pipeline
is similar: a projection mapping low-level features to the semantic
representation is learned from the auxiliary dataset by either classification
or regression models and applied directly to map each instance into
the same semantic representation space where a zero-shot classifier
is used to recognise the unseen target class instances with a single
known `prototype' of each target class. In this paper we discuss two
related lines of work improving the conventional approach: exploiting
transductive learning ZSL, and generalising ZSL to the multi-label
case.

\section{Transductive multi-class zero-shot learning }

Two inherent problems exist in the conventional ZSL formulation. (1)
projection domain shift problem: Since the two datasets have different
and potentially unrelated classes, the underlying data distributions
of the classes differ, so do the `ideal' projection functions between
the low-level feature space and the semantic spaces. Therefore, using
the projection functions learned from the auxiliary dataset without
any adaptation to the target dataset causes an unknown shift/bias.
This is illustrated in Fig.~\ref{fig:project_domain_shift_tsne}(A),
where both Zebra (auxiliary) and Pig (target) classes in AwA dataset
share the same `hasTail' semantic attribute, yet with different visual
appearance of their tails. Similarly, many other attributes of Pig
are visually different from the corresponding attributes in the auxiliary
classes. Figure~\ref{fig:project_domain_shift_tsne}(A-b) illustrates
the projection domain shift problem by plotting an 85D attribute space
representation of image feature projections and class prototypes:
a large discrepancy exists between the Pig prototype and the projections
of its class member instances, but not for Zebra. This discrepancy
inherently degrades the effectiveness of ZSL for class Pig. This problem
has neither been identified nor addressed in the zero-shot learning
literature. (2) \emph{Prototype sparsity problem}: for each target
class, we only have a single prototype which is insufficient to fully
describe the class distribution. As shown in Figs.~\ref{fig:project_domain_shift_tsne}(B-b)
and (B-c), there often exist large intra-class variations and inter-class
similarities. Consequently, even if the single prototype is centred
among its class members in the semantic representation space, existing
ZSL classifiers still struggle to assign the correct class labels
to these highly overlapped data points -- one prototype per class
simply is not enough to model the intra-class variability. This problem
has never been explicitly identified although a partial solution exists~\cite{transferlearningNIPS}.

In addition to these inherent problems, conventional approaches to
ZSL are also limited in \textbf{exploiting multiple intermediate semantic
spaces/views}, each of which may contain complementary information
-- they are useful in distinguishing different classes in different
ways. In particular, while both visual attributes \cite{lampert2009zeroshot_dat,farhadi2009attrib_describe,liu2011action_attrib,yanweiPAMIlatentattrib}
and linguistic semantic representations such as word vectors \cite{wordvectorICLR,DeviseNIPS13,RichardNIPS13}
have been independently exploited successfully, multiple semantic
`views' have not been exploited. This is challenging because they
are often of very different dimensions and types and each suffers
from different domain shift effects discussed above. Moreover, the
exploitation has to be transductive for zero-shot learning as only
unlabelled data are available for the target classes.

In our work \cite{transductiveEmbeddingJournal,yanweiembedding},
we propose to solve the projection domain shift problem using a transductive
multi-view embedding framework. Under our framework, each unlabelled
instance from the target dataset is represented by multiple views:
its low-level feature view and its (biased) projections in multiple
semantic spaces (visual attribute space and word space in this work).
We introduce a multi-view semantic space alignment process to correlate
different semantic views and the low-level feature view by projecting
them onto a latent embedding space learned using multi-view Canonical
Correlation Analysis (CCA)~\cite{multiviewCCAIJCV}.
Learning this new embedding space is to transductively (using the
unlabelled target data) aligns the semantic views with each other,
and with the low-level feature view, thus rectifying the projection
domain shift problem.  Even with the proposed transductive multi-view
embedding framework, the prototype sparsity problem remains -- instead
of one prototype per class, a handful are now available, but they
are still sparse. Our solution to this problem is to explore the manifold
structure of the data distributions of different views projected onto
the same embedding space via label propagation on a graph. To this
end, we introduce novel transductive multi-view Bayesian label propagation
(TMV-BLP) algorithm for recognition in \cite{yanweiembedding} which
combines multiple graphs by Bayesian model averaging in the embedding
space. In our journal version \cite{transductiveEmbeddingJournal},
we further introduce a novel transductive multi-view hypergraph label
propagation (TMV-HLP) algorithm for recognition. The core of our TMV-HLP
algorithm is a new distributed representation of graph structure termed
heterogeneous hypergraph. Instead of constructing hypergraphs independently
in different views (i.e.~homogeneous hypergraphs),
data points in different views are combined to compute multi-view
heterogeneous hypergraphs. This allows us to exploit the complementarity
of different semantic and low-level feature views, as well as the
manifold structure of the target data to compensate for the impoverished
supervision available in the form of the sparse prototypes. Zero-shot
learning is then performed by semi-supervised label propagation from
the prototypes to the target data points within and across the graphs.
Some results are shown in Tab.~\ref{tab:Comparison-with-stateofart}
and Fig. \ref{fig:project_domain_shift_tsne}(B).

\selectlanguage{english}%
\begin{table}
\begin{centering}
\scalebox{0.6}{%
\begin{tabular}{c|c|c|c|c|c|c}
\hline 
{\small{}Approach } & \multicolumn{1}{c|}{{\small{}AwA ($\mathcal{H}$ \cite{lampert2009zeroshot_dat})}} & {\small{}AwA ($\mathcal{O}$) } & {\small{}AwA $\left(\mathcal{O},\mathcal{D}\right)$ } & {\small{}USAA } & {\small{}CUB ($\mathcal{O}$) } & {\small{}CUB ($\mathcal{F}$) }\tabularnewline
\hline 
{\small{}DAP } & {\small{}40.5(\cite{lampert2009zeroshot_dat}) / 41.4(}\textcolor{black}{\small{}\cite{lampert13AwAPAMI})
/ 38.4{*}}{\small{} } & {\small{}51.0{*} } & {\small{}57.1{*} } & {\small{}33.2(\cite{yanweiPAMIlatentattrib,fu2012attribsocial}) /
35.2{*} } & {\small{}26.2{*} } & {\small{}9.1{*}}\tabularnewline
{\small{}IAP } & {\small{}27.8(\cite{lampert2009zeroshot_dat}) / 42.2(}\textcolor{black}{\small{}\cite{lampert13AwAPAMI})}{\small{} } & {\small{}-- } & {\small{}-- } & {\small{}-- } & {\small{}-- } & {\small{}--}\tabularnewline
{\small{}M2LATM \cite{yanweiPAMIlatentattrib} } & {\small{}41.3 } & {\small{}-- } & {\small{}-- } & {\small{}41.9 } & {\small{}-- } & {\small{}--}\tabularnewline
{\small{}ALE/HLE/AHLE \cite{labelembeddingcvpr13} } & {\small{}37.4/39.0/43.5 } & {\small{}-- } & {\small{}-- } & {\small{}-- } & {\small{}-- } & {\small{}18.0}\tabularnewline
{\small{}Mo/Ma/O/D \cite{marcuswhathelps} } & {\small{}27.0 / 23.6 / 33.0 / 35.7 } & {\small{}-- } & {\small{}-- } & {\small{}-- } & {\small{}-- } & {\small{}--}\tabularnewline
{\small{}PST \cite{transferlearningNIPS} } & {\small{}42.7 } & {\small{}54.1{*} } & {\small{}62.9{*} } & {\small{}36.2{*} } & {\small{}38.3{*} } & {\small{}13.2{*}}\tabularnewline
{\small{}\cite{unifiedProbabICCV13} } & {\small{}43.4 } & {\small{}-- } & {\small{}-- } & {\small{}-- } & {\small{}-- } & {\small{}--}\tabularnewline
{\small{}\cite{Yucatergorylevel} } & {\small{}48.3{*}{*} } & {\small{}-- } & {\small{}-- } & {\small{}-- } & {\small{}-- } & {\small{}--}\tabularnewline
\hline 
{\small{}TMV-BLP\cite{yanweiembedding}} & {\small{}47.1} & {\small{}--} & {\small{}--} & {\small{}47.8} & {\small{}--} & {\small{}--}\tabularnewline
\hline 
{\small{}TMV-HLP \cite{transductiveEmbeddingJournal}} & \textbf{\small{}49.0}{\small{} } & \textbf{\small{}73.5}{\small{} } & \textbf{\small{}80.5}{\small{} } & \textbf{\small{}50.4}{\small{} } & \textbf{\small{}\-\-\-\-\-\-47.9}{\small{} } & \textbf{\small{}19.5}\tabularnewline
\hline 
\end{tabular}}
\par\end{centering}

\noindent \protect\caption{\scriptsize \label{tab:Comparison-with-stateofart}Comparison with
the state-of-the-art on zero-shot learning on AwA, USAA and CUB. Features
$\mathcal{H}$, $\mathcal{O}$ and $\mathcal{F}$ represent hand-crafted,
OverFeat and Fisher Vector respectively. Mo, Ma, O and D represent
the highest results in the mined object class-attribute associations,
mined attributes, objectness as attributes and direct similarity methods
used in \cite{marcuswhathelps} respectively. `--': no result reported.
{*}: our implementation. {*}{*}: requires additional human interventions.}
\end{table}

\selectlanguage{british}%
\vspace{-1cm}

\section{Transductive multi-label zero-shot learning}

Many real-world data are intrinsically multi-label. For example, an image on Flickr often contains multiple
objects with cluttered background, thus requiring more than one label to describe its content.
And different labels are often correlated (e.g. cows often appear on grass).
In order to better predict these labels given an image, the label correlation must be modelled: for $n$ labels, there are $2^n$ possible multi-label combinations and to collect sufficient training samples for each combination to learn the correlations of labels is infeasible. More fundamentally, existing multi-class ZSL algorithms cannot model any such correlation as no labeled examples are available in this setting.

 We propose a novel framework for multi-label zero-shot learning~\cite{yanweiBMVC}. Given an auxiliary dataset containing labelled images, and a target dataset \emph{multi-labelled} with unseen classes (i.e.~none of the labels appear in the training set), we aim to learn a zero-shot model that performs multi-label classification on the test set with unseen labels. Zero-shot transfer is achieved using an intermediate semantic representation in the form of the skip-gram word vectors  \cite{distributedword2vec2013NIPS} which allows
vector-oriented reasoning. For example,
$Vec(`Moscow')$ is closer to $Vec(`Russia')+Vec(`capital')$ than
$Vec(`Russia')$ or $Vec(`capital')$ only. This property will enable zero-shot multi-label
prediction by enabling synthesis  of multi-label prototypes  in the semantic word space. 

\begin{figure}
\begin{centering}
\begin{tabular}{cc}
\includegraphics[scale=0.17]{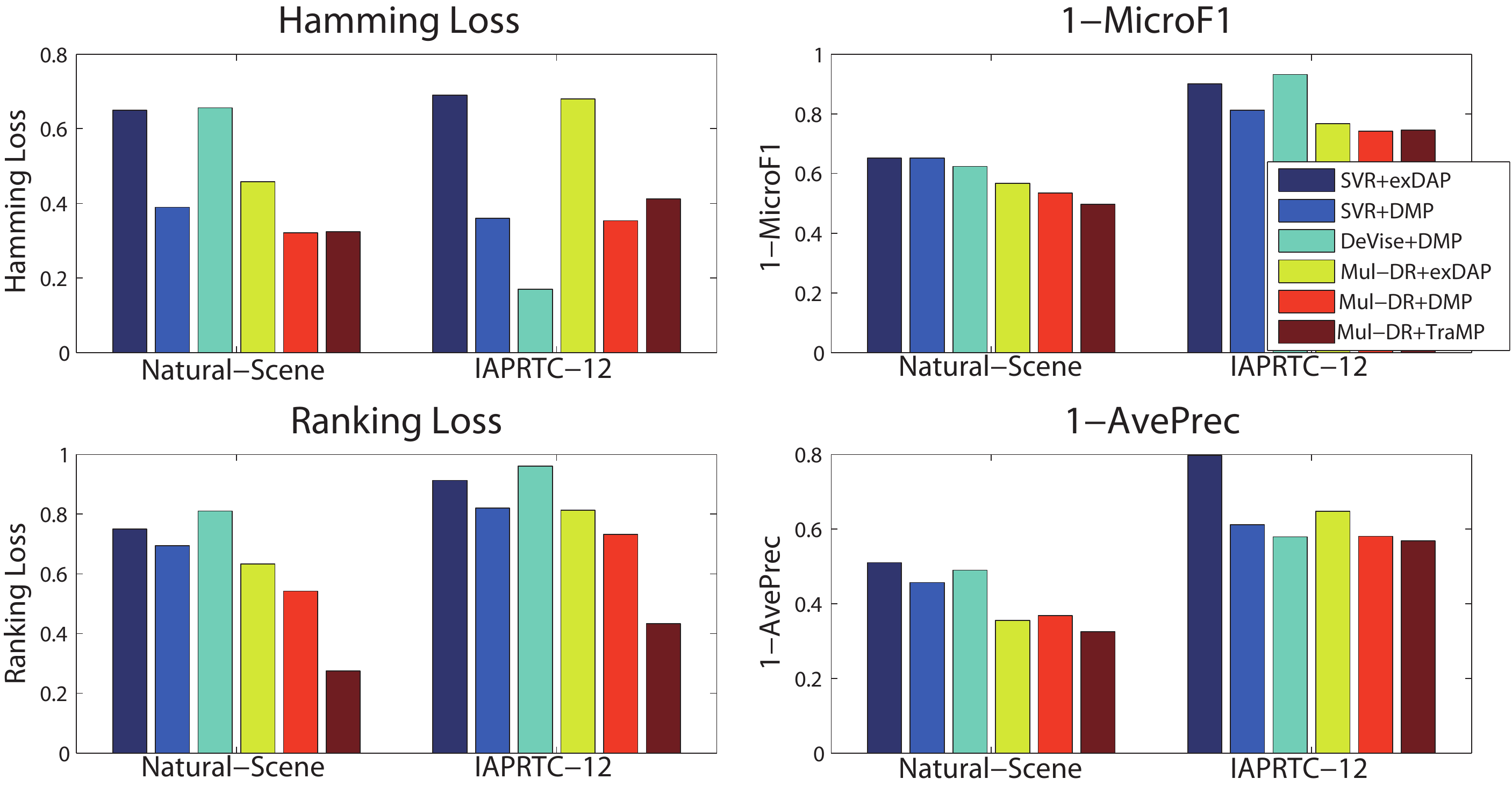} &
\includegraphics[scale=0.5]{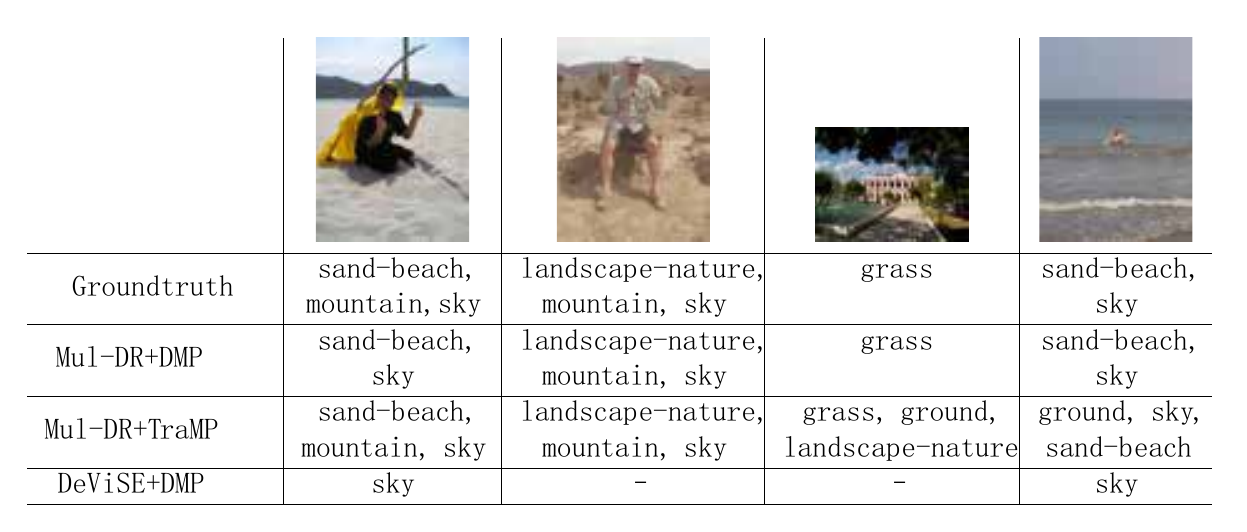} \\
(A) &(B)
\end{tabular}
\par\end{centering}

\protect\caption{\label{fig:Comparing-methods-ofMLdataset}\scriptsize{(A) Comparing different zero-shot multi-label classification methods on Natural
Scene and IAPRTC-12.So smaller values for all metrics are preferred. (B) Examples of  ML-ZSL predictions on IAPRTC-12. Top 8 most frequent labels of landscape-nature branch are considered.}}

\end{figure}

Our framework has two main
components: multi-output deep regression (Mul-DR) and  zero-shot multi-label
prediction (ZS-MLP). Mul-DR is a 9 layer neural
network that exploits convolutional neural network
(CNN) layers, and includes two multi-output regression layers as the final layers. It learns
from auxiliary data the mapping from raw image pixels to a linguistic representation
defined by the skip-gram language model \cite{distributedword2vec2013NIPS}. With \textbf{Mul-DR}, each test image is now projected into the semantic word space where the unseen labels and their combinations can be represented as data points without the need to collect any visual data.
\textbf{ZS-MLP} addresses the multi-label ZSL problem in this semantic word space by exploiting the property that label combinations can be synthesised.
We exhaustively synthesise the power set of all possible prototypes
(i.e., combinations of multi-labels) to be treated as if they were a  set of labelled
instances in the space. With this synthetic dataset, we are able to
 propose two new multi-label algorithms -- direct multi-label zero-shot
 prediction (DMP) and transductive multi-label zero-shot prediction
(TraMP). However, Mul-DR is learned using the auxiliary classes/labels, so it may not generalise well to the unseen classes/labels (projection domain shift problem, as discussed in the previous section). To overcome this problem, we further exploit self-training to adapt Mul-DR to the test classes to improve its generalisation capability.
 The experimental results on Natural Scene and IAPRTC-12 in Fig \ref{fig:Comparing-methods-ofMLdataset} show the efficacy of our framework for multi-label ZSL over a variety of baselines. For more details, please read our paper \cite{yanweiBMVC}.

\clearpage

\bibliographystyle{abbrv}
\bibliography{PhD_thesis_ref}

\end{document}